\documentclass[10pt,conference]{IEEEtran}
\IEEEoverridecommandlockouts

\usepackage{cite}
\usepackage[utf8]{inputenc}
\usepackage{amsmath,amssymb,amsfonts}
\usepackage{algorithmic}
\usepackage{graphicx}
\usepackage{textcomp}
\usepackage{xcolor}
\usepackage{balance}
\usepackage{makecell}
\usepackage{pifont}
\usepackage{url}

\usepackage{hyperref}
\DeclareRobustCommand{\myurl}[1]{\url{#1}}

\def\BibTeX{{\rm B\kern-.05em{\sc i\kern-.025em b}\kern-.08em T\kern-.1667em\lower.7ex\hbox{E}\kern-.125emX}}

\begin{document}

\def\Rbb{{\mathbb{R}}}
\def\Cbb{{\mathbb{C}}}
\def\Fbf{{\mathbf F}}
\def\Gbf{{\mathbf G}}
\def\Hbf{{\mathbf H}}
\def\Ubf{{\mathbf U}}
\def\Vbf{{\mathbf V}}
\def\Ybf{{\mathbf Y}}
\def\Xbf{{\mathbf X}}
\def\Ibf{{\mathbf I}}
\def\Sigmabf{\mathbf{\Sigma}}
\def\Pcal{{\mathcal P}}
\def\Ccal{{\mathcal C}}
\def\Ncal{{\mathcal N}}

\def\fbf{{\mathbf f}}
\def\vbf{{\mathbf v}}
\def\gbf{{\mathbf g}}
\def\xbf{{\mathbf x}}
\def\sbf{{\mathbf s}}
\def\cbf{{\mathbf c}}
\def\ybf{{\mathbf y}}
\def\zbf{{\mathbf z}}
\def\st{{\mathbf{s}_{\mathrm{T}}}}
\def\sr{{\mathbf{s}_{\mathrm{R}}}}
\def\Nt{{{N}_{\mathrm{T}}}}
\def\Nr{{{N}_{\mathrm{R}}}}

\def\snrDeepJSCC{{\text{SNR}_\text{DeepJSCC}}}
\def\snralign{{\text{SNR}_\text{Align}}}

\title{Semantic Channel Equalization Strategies for \\Deep Joint Source-Channel Coding 
\thanks{This work was supported by the European Union under the Italian National Recovery and Resilience Plan (NRRP) of NextGenerationEU, partnership on “Telecommunications of the Future” (PE00000001 - program “RESTART”), by the Next Generation EU, Mission I.3.3 PNRR Scholarships for Innovative PhD Programmes Addressing Business Innovation Needs, Component 2, CUP 352: B53C22001970004, and by the SNS JU project 6G-GOALS under the EU’s Horizon program Grant Agreement No 101139232.\\
All source code is available on GitHub at the following link:\\\url{https://github.com/SPAICOM/DJSCC-Semantic-Equalization}}
}

\author{Lorenzo Pannacci, Simone Fiorellino$^{1,2}$, Mario Edoardo Pandolfo$^{1,2}$,\\ Emilio Calvanese Strinati$^{3}$, and Paolo Di Lorenzo$^{2,4}$ \smallskip\\
$^1$ DIAG Department, Sapienza University of Rome, via Ariosto 25, Rome, Italy.\\
$^2$ Consorzio Nazionale Interuniversitario per le Telecomunicazioni (CNIT), Parma, Italy.\\
$^3$ CEA Leti, University Grenoble Alpes, 38000, Grenoble, France.\\
$^4$ DIET Department, Sapienza University of Rome, Via Eudossiana 18, Rome, Italy. \smallskip\\
E-mail:  \{simone.fiorellino,marioedoardo.pandolfo,paolo.dilorenzo\}@uniroma1.it,\\ \{emilio.calvanese-strinati\}@cea.fr, lorenzo.pannacci.01@gmail.com. 
}

\maketitle

\begin{abstract}
Deep joint source–channel coding (DeepJSCC) has emerged as a powerful paradigm for end-to-end semantic communications, jointly learning to compress and protect task-relevant features over noisy channels. However, existing DeepJSCC schemes assume a shared latent space at transmitter (TX) and receiver (RX)—an assumption that fails in multi-vendor deployments where encoders and decoders cannot be co-trained. This mismatch introduces “semantic noise”, degrading reconstruction quality and downstream task performance. In this paper, we systematize and evaluate methods for \emph{semantic channel equalization} for DeepJSCC, introducing an additional processing stage that aligns heterogeneous latent spaces under both physical and semantic impairments. We investigate three classes of aligners: (i) linear maps, which admit closed-form solutions; (ii) lightweight neural networks, offering greater expressiveness; and (iii) a Parseval-frame equalizer, which operates in zero-shot mode without the need for training. Through extensive experiments on image reconstruction over AWGN and fading channels, we quantify trade-offs among complexity, data efficiency, and fidelity, providing guidelines for deploying DeepJSCC in heterogeneous AI-native wireless networks.
\end{abstract}

\begin{IEEEkeywords}
Semantic channel equalization, DeepJSCC, latent spaces, AI-native communications, semantics.
\end{IEEEkeywords}

\section{Introduction}


Modern communication systems typically use a layered pipeline: source coding (e.g., JPEG, WebP) reduces redundancy, followed by channel coding (e.g., LDPC) to protect the bitstream before modulation. This separation, based on Shannon’s Separation Theorem \cite{shannon1998mathematical}, is optimal under infinite block lengths and complexity, but assumes the physical layer need not consider how data is used. While this modularity has supported generations of mobile standards, it struggles under the tight latency, bandwidth, and energy constraints of emerging applications like IoT and autonomous driving. In such cases, joint source–channel coding (JSCC), which maps source data directly to channel symbols, can offer better performance \cite{gunduz2024joint}. In particular, the DeepJSCC paradigm parametrizes both encoder and decoder as deep neural networks, forming an autoencoder where channel noise is explicitly added to the transmitted latent representations during end-to-end training. 
Thanks to the flexibility of its loss architecture, DeepJSCC not only learns to compress the most task-relevant (semantic) features, but also to protect them against physical channel impairments. DeepJSCC has been studied for wireless transmission of text \cite{farsad2018text} and images \cite{bourtsoulatze2019deep,erdemir2023generative}, for MIMO channels \cite{bian2023space,wu2024deep}, for privacy-preserving transmission \cite{letafati2024deep}, and in schemes with feedback \cite{kurka2020DJSCC}. 
This unified view naturally supports the vision of semantic communications \cite{xu2023deep}, where the transmitted payload conveys just the relevant features for the downstream task, enabling consistent spectral-efficiency gains \cite{strinati20216g,strinati2024goal}.

Most existing DeepJSCC frameworks assume a \textit{shared} latent representation between the transmitter and receiver, relying on the premise that both neural networks at transmitter and receiver are trained jointly in an end-to-end manner. Under this assumption, \textit{semantic noise} \cite{luo2022semantic,getu2023semantic}---which stems from mismatches in logic, interpretation, or knowledge between AI-native devices---is effectively absent. 
However, in practical deployments---particularly those involving different vendors who are unwilling or unable to share training data, model architectures, or other proprietary assets---heterogeneous encoder–decoder pairs are often unavoidable. This results in mismatches between the latent spaces of AI-native devices, introducing \textit{semantic noise}. To enable effective communication in such cases, an additional processing step referred to as \textit{semantic channel equalization} is required. Among the strategies available in the literature, \textit{relative representations} (RRs) have emerged as a promising approach for zero-shot latent-space communication \cite{moschella2022relative}, and have been successfully applied to goal-oriented network optimization \cite{fiorellino2024dynamic}. Further developments have enhanced their application by eliminating the need for decoder retraining \cite{maiorca2024latent}, using a frame-based semantic channel equalizer \cite{fiorellino2025frame}, and enabling operation with minimal data sharing \cite{huttebraucker2024relative}. Notably, in \cite{pandolfo2025mimo}, semantic channel equalization is integrated directly into a MIMO system design---allowing joint optimization of semantic compression and latent-space alignment---while another recent work embeds reconfigurable intelligent surfaces (RISs) into the process \cite{huttebraucker2025ris}. These works show that physical and semantic impairments can be jointly addressed within a unified framework.

Despite these advancements, none of the existing semantic equalization approaches have addressed the challenges specific to DeepJSCC, nor have they investigated the practical implementation of semantic equalizers in such systems. This paper aims to fill this gap by proposing and evaluating semantic channel equalization techniques explicitly designed for DeepJSCC.
We introduce and evaluate three classes of aligners—linear transformations, lightweight neural networks (based on multi-layer perceptron or convolutional architectures), and a zero-shot Parseval-frame operator—analyzing their respective trade-offs in terms of computational complexity, data efficiency, robustness, and scalability. Through extensive image reconstruction experiments over AWGN and fading channels, we show that unaligned DeepJSCC pairs fail to recover semantically meaningful images in the presence of semantic noise. These findings establish alignment as a critical component for effective DeepJSCC,  offering practical deployment guidelines for multi-vendor, AI-native communications.

\section{System Model}
\label{sec:system_model}

\begin{figure}[t]
    \centering
    \includegraphics[width=\columnwidth, trim=0bp 0bp 0bp 0bp, clip]{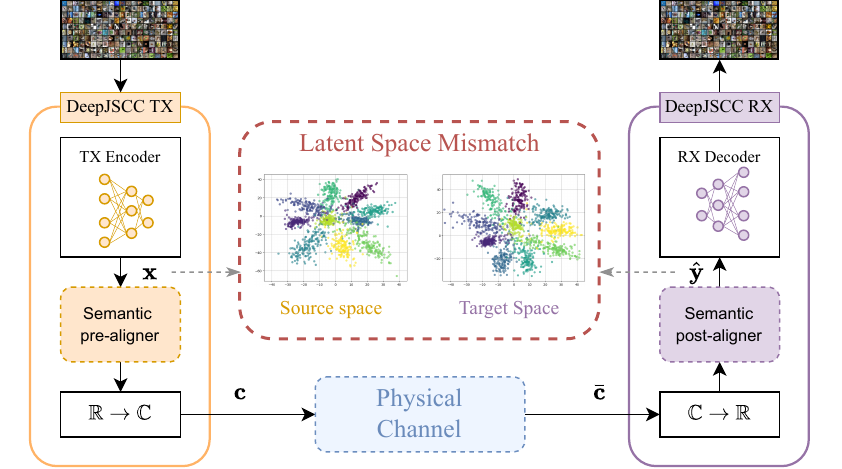}
    \caption{Scheme of the proposed DeepJSCC model with semantic equalization.}
    \label{fig:system_model}
\end{figure}

As illustrated in Fig. \ref{fig:system_model}, we consider a point-to-point link between two \emph{heterogeneous} agents, TX and RX, performing end-to-end joint source–channel coding (JSCC), with the encoder at TX and the decoder at RX. The two sides may utilize different network architectures and/or training strategies. Let $\mathbf{u}\in\mathbb{R}^{n}$ denote the source sample (e.g., a vectorized image). The TX transforms $\mathbf{u}$ into a $d$-dimensional real latent vector $\mathbf{x} \in \mathbb{R}^{d}$ via a learned mapping implemented with a CNN-based architecture \cite{bourtsoulatze2019deep}, enforcing an average power constraint $\,\mathbb{E} \left[\|\xbf\|^{2}\right]\le P_T$,
where $P_T$ is the transmission power. 
The collection of all such vectors forms the \emph{TX latent space}. On the other side, the RX has been trained under a different encoding scheme, or training process, and expects a different latent vector $\ybf \in \mathbb{R}^{m}$ to be able to understand and recover $\mathbf{u}$.

Semantic alignment can be implemented as a set of transformations applied entirely at the TX, entirely at the receiver RX, or split between the two. In the general case (i.e., split between the two), assuming $d$ is even, we first apply a norm-preserving \emph{semantic pre-aligner}, $\gbf:\Rbb^d\to\Rbb^{2k}$, that converts $\xbf$ into a $2k$-dimensional real vector. We then pair its entries to form the complex vector $\cbf \in \mathbb{C}^{k}$, which is more suitable for radio-frequency transmission. Let $\psi_{\mathbb{R}^{2k}\rightarrow \mathbb{C}^k}$ denote such real-to-complex mapping. Following the DeepJSCC literature, we refer to $n$ as the \emph{source bandwidth} and to $k$ as the \emph{channel bandwidth}, and define the \textit{bandwidth ratio} $\rho \triangleq {k}/{n}$. We assume that transmission takes place over a flat-fading Rayleigh channel in the presence of AWGN:
\begin{align}
    \bar{\cbf} = h\,\cbf + \mathbf{v},\quad\hbox{with} \quad
\cbf = \psi_{\mathbb{R}^{2k} \rightarrow \mathbb{C}^k}\bigl(\gbf(\xbf)\bigr),
\end{align}
where $h \in \mathbb{C}$ is a complex fading coefficient modeled as a zero-mean complex Gaussian representing the fading effect, and $\vbf \sim \mathcal{CN}(\mathbf{0}, \sigma_v^2 \mathbf{I}_k)$ is circularly symmetric complex Gaussian noise. Both the TX and RX DeepJSCCs are trained at a specific SNR, denoted $\mathrm{SNR}_{\mathrm{DeepJSCC}}$. At the RX side, $\bar{\cbf}$ is transformed back into a real-valued vector $\bar{\xbf}\in\Rbb^{d}$, and a \textit{semantic post-aligner} $\fbf:\Rbb^d\to\Rbb^m$ maps this vector into the latent space expected by the RX decoder, i.e., $\fbf(\bar{\xbf})\mapsto\hat{\ybf}$. Lastly, the RX decoder produces the task output from $\hat{\ybf}$.

To enable the optimization of the semantic aligner, we introduce the concept of \textit{semantic pilots}—pairs of source and target samples used to learn the alignment function $\mathbf{f}$. We denote the set of semantic pilots by $\mathcal{P}$, i.e., a collection of $N$ labeled pairs $\mathcal{P} = \bigl\{(\xbf^{(i)}, \ybf^{(i)})\bigr\}_{i=1}^{N}$,
where each $\mathbf{x}^{(i)}$ is the transmitted latent vector and $\mathbf{y}^{(i)}$ the corresponding ground-truth output. In practice, semantic pilots can be obtained through a supervised calibration phase, during which both the transmitter and receiver have access to a shared dataset of sample pairs. This dataset can be public, ensuring that neither agent is required to disclose private data. Moreover, since explicit labels are not strictly necessary, the data collection process remains lightweight and low in complexity.

For notational simplicity, we will omit the explicit real-to-complex mapping $\psi_{\mathbb{R}^{2k} \rightarrow \mathbb{C}^k}$ (and its inverse) in what follows, treating the conversion as implicit whenever a vector enters or exits the physical channel.
Once the set $\mathcal{P}$ of semantic pilots is available, the goal is to adjust the aligner $\mathbf{f}$ so as to minimize a semantic distance $d_{\mathrm{sem}}\bigl(\mathbf{y},\hat{\mathbf{y}}\bigr)$ over $\mathcal{P}$, thus aligning the TX and RX latent spaces in the presence of semantic and/or physical channel noise: 
\begin{align}\label{pr:general}
    \min_{\fbf}\;\mathbb{E}[ d_{\mathrm{sem}}\bigl(\mathbf{y},\;\fbf(h\gbf(\xbf) + \mathbf{v})\bigr)].
\end{align}
To optimize alignment under the intended channel conditions, the semantic aligner itself is trained at a potentially different SNR, denoted $\mathrm{SNR}_{\mathrm{Align}}$.

\section{Semantic Equalization Strategies}
\label{sec:methods}

This section introduces three semantic channel equalization strategies tailored for DeepJSCC: linear, neural, and zero-shot Parseval frame equalizers, each balancing trade-offs between robustness, data efficiency, and deployment complexity.

\subsection{Linear Semantic Equalizer}

The \emph{linear semantic equalizer} implements only the semantic \emph{post}–aligner, learning a single linear map that projects the received TX latent vectors onto the RX latent space. To this aim, we model the aligner $\fbf(\cdot)$ by a matrix $\Fbf\in\mathbb{R}^{m \times d}$. The linear semantic aligner is obtained by solving
\begin{align}\label{pr:lse_base}
\min_{\Fbf}&\quad \mathbb{E}\left[||\ybf-\Fbf\cbf||^2_2\right] 
\end{align}
where $\mathbb{E}$ denotes the expected value over the noise and data distributions, and we assumed mean squared error as a simple example of semantic distance $d_{\mathrm{sem}}(\cdot,\cdot)$.  Leveraging the zero-mean property of the noise, the objective function of \eqref{pr:lse_base} can be approximated with the empirical average over data points and fading coefficients:
\begin{align}
\min_{\Fbf}&\quad \frac{1}{N}\sum_{i=1}^{N}||\ybf^{(i)}-\Fbf \tilde{\xbf}^{(i)}||^2_2 + \text{tr}(\Fbf\tilde\Sigma_\vbf\Fbf^H)
\end{align}
with $\tilde\Sigma_\vbf$ denoting the real noise covariance matrix, and $\tilde{\xbf}^{(i)}$ is used to denote the transmitted latent vector scaled by the corresponding fading coefficient. 
Finally, stacking the semantic pilots $\Pcal=\{(\xbf^{(i)},\ybf^{(i)})\}_{i=1}^{N}$ into the matrix $\Ybf\in\mathbb{R}^{m\times N}$, and defining the matrix of scaled latent vectors $\tilde{\Xbf}=[\tilde{\xbf}^{(1)},\ldots,\tilde{\xbf}^{(N)}]\in\mathbb{R}^{d\times N}$, the optimization problem can be compactly written as:
\begin{align}\label{pr:lse_matrix_form}
\min_{\Fbf}&\quad \frac{1}{N}||\Ybf-\Fbf\tilde\Xbf ||^2_F + \text{tr}(\Fbf\tilde\Sigma_\vbf\Fbf^H).
\end{align}
Problem \eqref{pr:lse_matrix_form} allows a closed form solution obtained by setting the gradient of the objective function w.r.t. $\Fbf^H$ to zero,
\begin{align}
    \Fbf^*=\Ybf\tilde\Xbf^H(\tilde\Xbf\tilde\Xbf^H+ N\Sigma_\vbf)^{-1}.
\end{align}
Linear Semantic equalizers are attractive for their simplicity and low complexity, although their performance is ultimately bounded by the expressiveness of a single matrix product.

\begin{figure*}[t]
    \centering
    \includegraphics[width=0.75\textwidth, trim=7bp 0bp 0bp 7bp, clip]{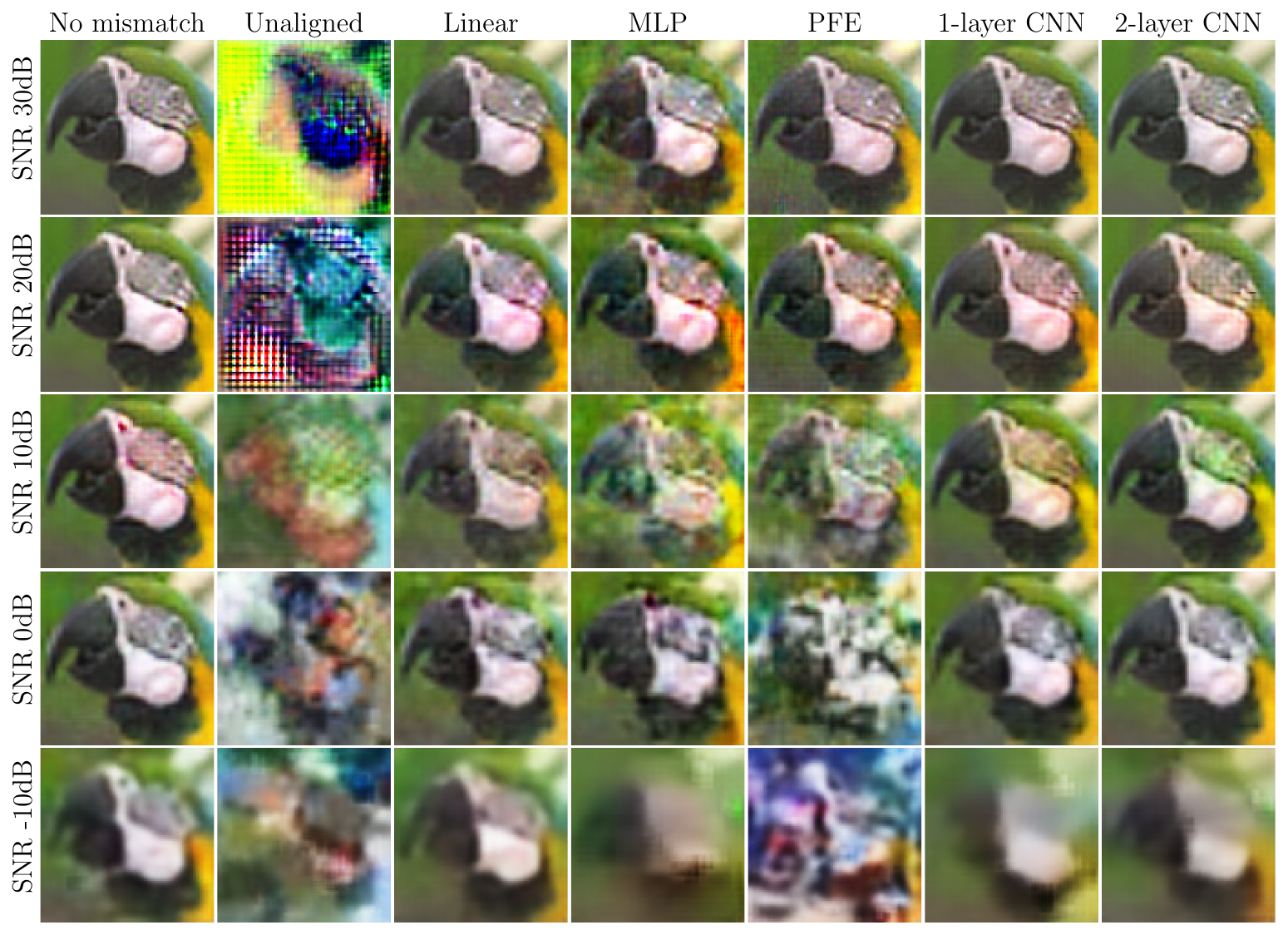}
    \caption{Reconstruction of a Kodak patch at DeepJSCC-trained SNRs \{$-10$, $0$, $10$, $20$, $30$\} dB, evaluated only under the presence of AWGN at the same SNR. Results are shown for no mismatch, mismatched without equalizer, and results with Linear, MLP, 2-layer CNN, 1-layer CNN, and PFE equalizers (all with $|\Pcal|= 10\,000$ pilots).}
    \label{fig:visual_psnr}
\end{figure*}

\subsection{Neural Semantic Equalizer}

As with the linear case, here we implement only a \emph{semantic post–aligner}. Linear equalizers typically admit closed-form or otherwise simpler optimization procedures, and their limited representational capacity often fails to capture complex relationships. To overcome this limitation, we introduce a more expressive alternative: the \textit{neural semantic equalizer}. We define the semantic aligner as a parametric neural function $\fbf_\theta:\Rbb^m \to \Rbb^d$, where $\theta$ denotes the trainable parameters. The network is trained by minimizing the expected semantic reconstruction loss over the noisy observations:
\begin{align}\label{pr:nse_base}
\min_{\theta}&\quad \mathbb{E}\left[||\ybf-\fbf_\theta\left(\cbf\right)||^2_2\right] 
\end{align}
During training, noise is added to the input in each forward pass, effectively simulating channel corruption and promoting robustness to physical-layer impairments. In our experiments, we consider two lightweight but expressive architectures. The first is a single-hidden-layer multilayer perceptron (MLP) with a PReLU activation, which already offers more flexibility than a linear map.  The second is a shallow convolutional network with two convolutional layers separated by a PReLU.  Because both the TX and RX encoders themselves are CNN-based, this convolutional aligner naturally exploits the local structure and weight sharing of their latent representations. Finally, we include a simpler variant formed by a single convolutional layer, which reduces exactly to a linear equalizer, providing a baseline for comparison.

\subsection{Parseval Frame Equalizer}
\label{sec:frame_method}
We now present a semantic channel equalizer that requires no joint training of semantic pilots and is channel-agnostic, originally introduced in \cite{fiorellino2025frame}. The TX and RX are equipped with an ordered reference set composed of input data points (e.g., images) denoted as $\mathcal{S}=\{\zbf_{1},\dots,\zbf_{|\mathcal{S}|}\}$, assumed to be pre-agreed and known to both parties.
Encoding these samples in their respective (private) latent spaces yields two ordered matrices  
$\tilde{\mathbf{G}}=[\xbf_1,\dots,\xbf_{2k}]^{ \top} \in \mathbb{R}^{2k\times d}$, the semantic pre-aligner, and  
$\tilde{\mathbf{F}}=[\ybf_1,\dots,\ybf_{2k}]^{ \top} \in \mathbb{R}^{2k\times m}$, the semantic post-aligner.  
The $n$-th row of each matrix corresponds to the same semantic concept at the two ends. To obtain well-conditioned operators, we normalize them as:
\begin{equation}
\mathbf{G}= \tilde{\mathbf{G}}\bigl(\tilde{\mathbf{G}}^{ H}\tilde{\mathbf{G}}\bigr)^{-1/2},
\qquad
\mathbf{F}= \tilde{\mathbf{F}}\bigl(\tilde{\mathbf{F}}^{ H}\tilde{\mathbf{F}}\bigr)^{-1/2}    
\end{equation}
yields $\mathbf{G}^{ H}\mathbf{G}= \mathbf{I}_{d}$ and  
$\mathbf{F}^{ H}\mathbf{F}= \mathbf{I}_{m}$.  For $2k\gg d$ (or $2k\gg m$), these operators form an overcomplete Parseval frame. Otherwise, under compression, they remain perfectly conditioned on their spanned subspaces.
At runtime, TX computes an analysis operation $\mathbf{c}= \mathbf{G}\mathbf{x}\in\mathbb{C}^{2k}$, which maps each latent feature to the agreed reference directions, and sends $k$ complex coefficients over the channel. At RX side a synthesis operation is computed, i.e.,  $\hat{\mathbf{y}}=\mathbf{F}^{\mathrm{H}}\mathbf{c}$, 
recovering a latent vector that is aligned concept-by-concept with its own decoder.

The resulting \emph{Parseval-Frame Equalizer} (PFE) operates in zero-shot mode and is numerically robust, requiring only the ordered sequence of the data points used for the operator composition, therefore avoiding the SPs transmission.

\section{Numerical Results}
\label{sec:numerical_results}

In this section, we evaluate the performance of the considered semantic equalization methods through numerical experiments.
We use the CIFAR-10 dataset \cite{krizhevsky2009learning}, which comprises $60\,000$ color images of size $32  \times  32$ across $10$ mutually exclusive classes. Following common practice, we partition the dataset into $50\,000$ training images and $10\,000$ test images; all images are upscaled to $96 \times 96$ pixels at inference time. All DeepJSCC models are trained as in \cite{bourtsoulatze2019deep} with a compression rate of $k/n = 1/6$.
Both TX and RX employ end-to-end DeepJSCC autoencoders with convolutional encoders and decoders.
Training is carried out over a flat-fading Rayleigh channel with unit variance, in the presence of AWGN, at a fixed SNR, denoted as $\text{SNR}_{\text{DeepJSCC}}$. To introduce semantic noise, the two models are trained with different random seeds, 42 for TX and 43 for RX. During evaluation, the latent code produced by the TX encoder is transmitted over the channel and decoded directly by the RX decoder. The quality of image reconstruction is measured in terms of peak signal-to-noise ratio (PSNR).

\begin{figure}[t]
    \centering    \includegraphics[width=0.83\columnwidth, trim=0bp 5bp 0bp 0bp, clip]{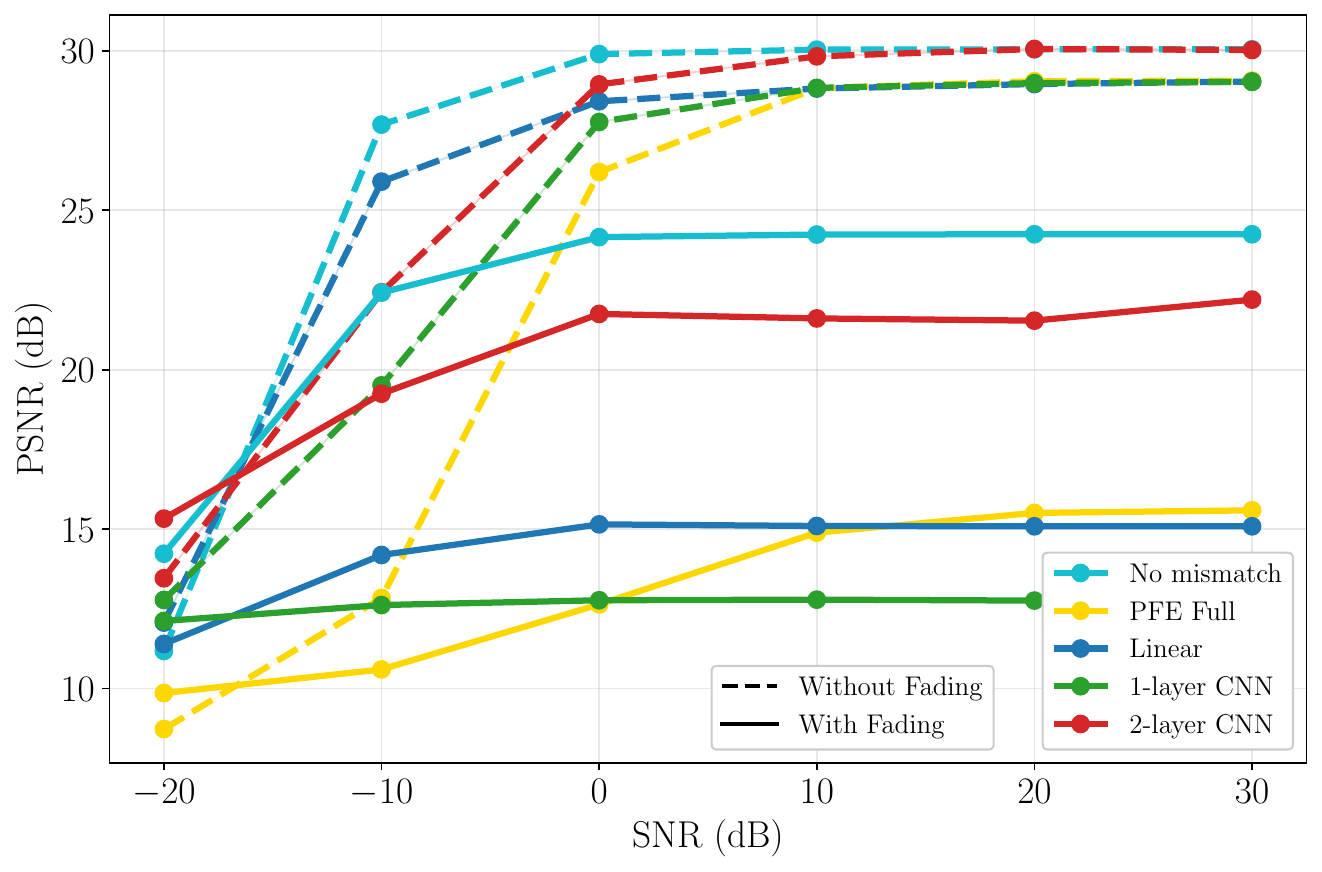}
    \caption{PSNR versus $\snralign$, with $\snrDeepJSCC\!=\!-10\,$dB and $|\Pcal|\!=\!10\,000$.}
    \label{fig::psnr_snr_10}
\end{figure}

\begin{figure}[t]
    \centering    \includegraphics[width=0.83\columnwidth, trim=0bp 0bp 0bp 0bp, clip]{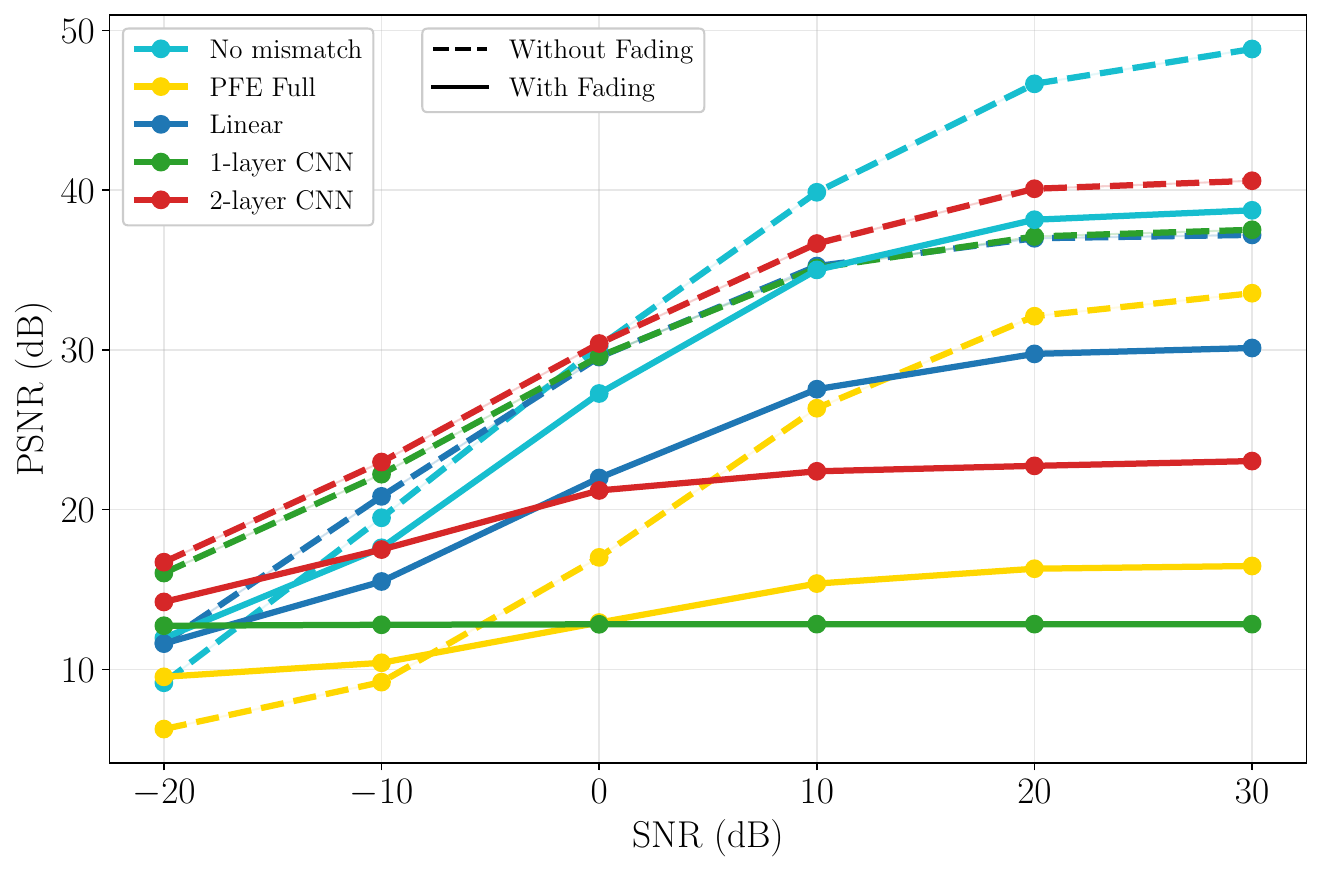}
    \caption{PSNR versus $\snralign$, with $\snrDeepJSCC=20\,$dB and $|\Pcal|=10\,000$.}
    \label{fig::psnr_snr_20}
\end{figure}

At the RX, we apply one of five semantic-equalization strategies to mitigate this mismatch: Linear, MLP, 1-layer CNN, 2-layer CNN, and PFE. PFE’s analysis operator $\Gbf$ projects the $d$-dimensional latent vector into $2k$ coefficients---one per reference sample---thereby inherently introducing an additional, dynamic level of compression. To isolate this effect, we distinguish two variants: PFE-full, which transmits $2k$ coefficients, matching the TX latent dimension; PFE, which transmits only as many coefficients as the number of semantic pilots, highlighting sample-efficiency trade-offs. Aligners are trained with varying amounts of semantic pilots. From the training set, a permutation is computed and subsets of different sizes are retrieved by taking the first indices, ensuring an incremental dataset.  
The neural semantic equalizers are trained using the Adam optimizer with adaptive early stopping, validation-based model selection and a batch size of 64. 10\% of the train set is dedicated to validation, and in the scenario of $|\Pcal| < 10$, the train loss is used instead of the validation loss. Consistently with the DeepJSCC setup, the optimization routine applies the same channel model described in Sec. \ref{sec:system_model}, with the SNR set to $\mathrm{SNR}_{\mathrm{Align}}$. Learning rate is $1e\!-\!4$ for the MLP aligner and $1e\!-\!3$ for the CNN-based ones. The MLP aligner training uses no weight decay, while the CNN-based models both use a weight decay of $0.001$. All convolutional layers have a kernel size of 5. Hidden dimensions are kept equal to input dimensions.
\\
For all results, we adopt a simplified channel model, where the fading coefficient is fixed to a deterministic unit value, so that only AWGN is present. For Figs. \ref{fig::psnr_snr_10} and \ref{fig::psnr_snr_20}, we also consider the full channel model of Sec. \ref{sec:system_model}, which includes both Rayleigh fading and AWGN.

Figure \ref{fig:visual_psnr} provides a qualitative comparison of the equalization methods on a single patch taken from an image from the Kodak Lossless True Color Image Suite \cite{kokakimagedataset}. Each row corresponds to a different SNR the underlying DeepJSCC autoencoders were trained on; $\mathrm{SNR}_{\mathrm{DeepJSCC}}\in\{-10,0,10,20,30\}\,\mathrm{dB}$, with the aligners being trained on the same SNR, i.e., $\mathrm{SNR}_{\mathrm{Align}} = \mathrm{SNR}_{\mathrm{DeepJSCC}}$. Each column shows one of the following cases: no semantic mismatch (TX and RX share the same DeepJSCC model), semantic mismatch without any aligner, and the five alignment strategies (Linear, MLP, two-layer CNN, single-layer CNN, PFE). All aligners were trained with $10\,000$ semantic pilots, and the random seed is set to 42 to ensure that each aligner uses the same pilots. As expected, at high $\mathrm{SNR}$ all methods yield reconstructions close to the non-mismatched case, whereas at low $\mathrm{SNR}$ only the linear equalizer preserves recognizable image details.
\begin{figure}[t]
    \centering    \includegraphics[width=0.83\columnwidth, trim=0bp 0bp 0bp 0bp, clip]{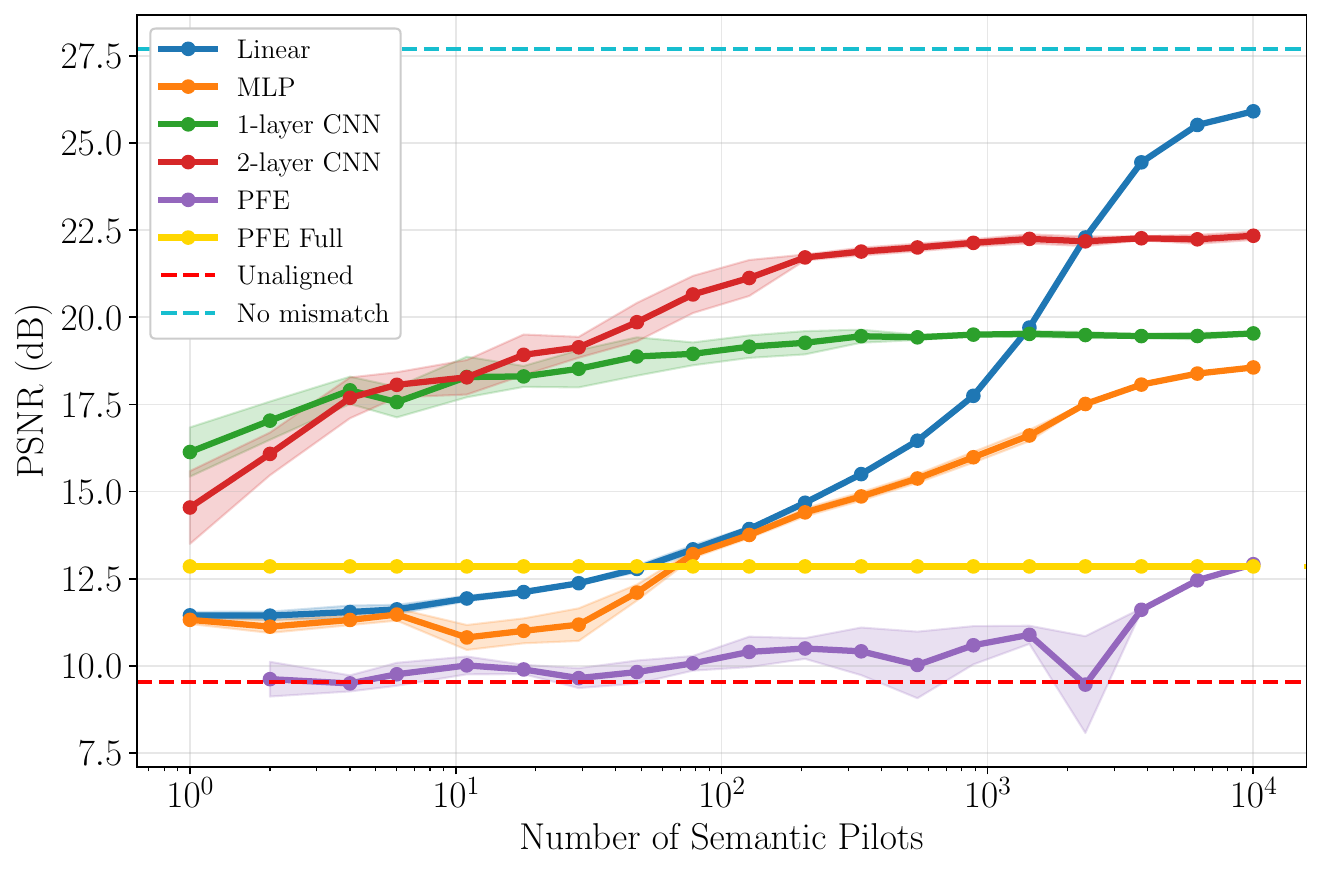}
    \caption{PSNR versus $|\Pcal|$, with both $\snralign$ and $\snrDeepJSCC$ fixed at $-10$ dB, considering only the AWGN component.}
    \label{fig::snr-10}
\end{figure}
\begin{figure}[t]
    \centering    \includegraphics[width=0.83\columnwidth, trim=0bp 0bp 0bp 0bp, clip]{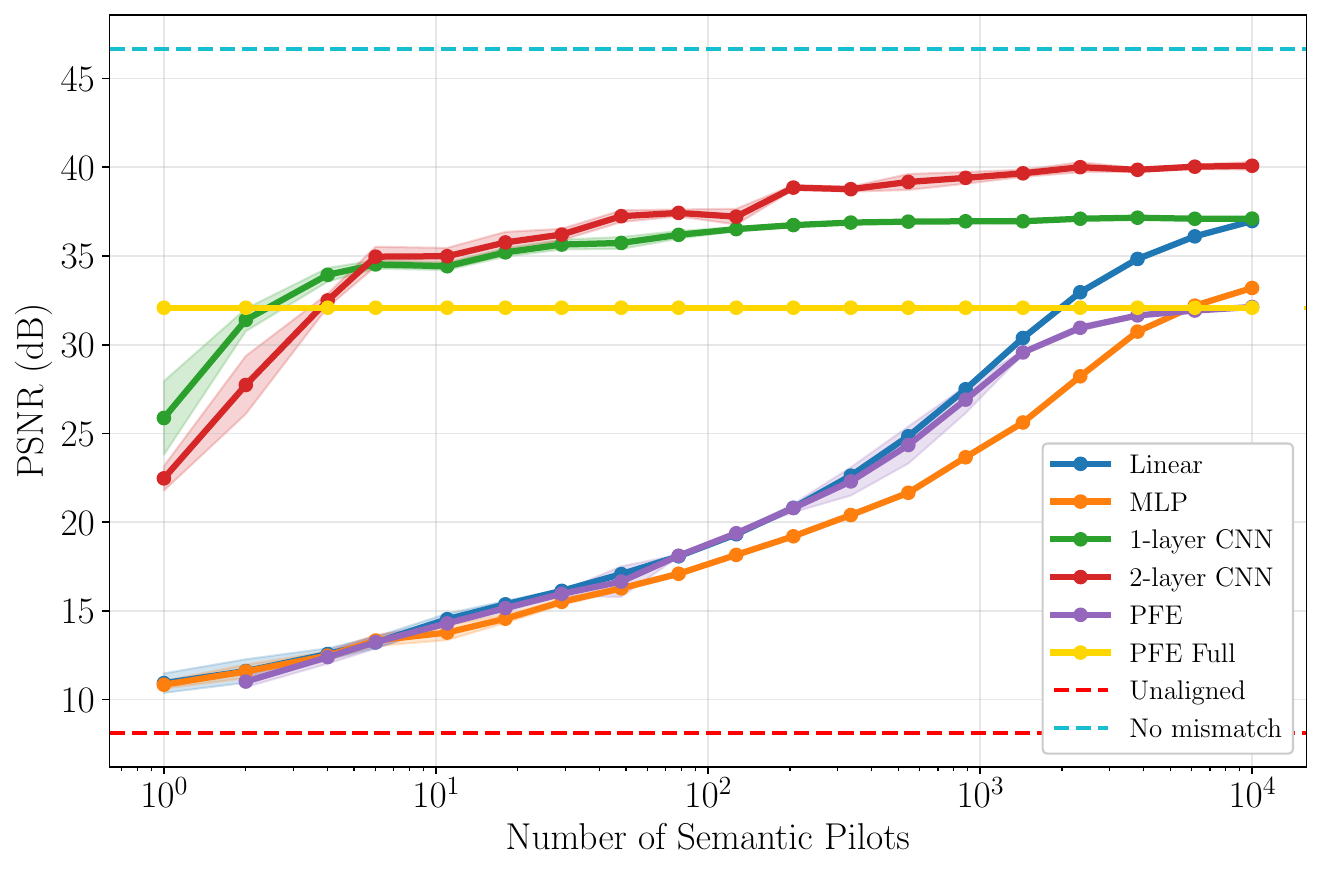}
    \caption{PSNR versus $|\Pcal|$, with both $\snralign$ and $\snrDeepJSCC$ fixed at $20$ dB, considering only the AWGN component.}
    \label{fig::snr20}
\end{figure}

Figures \ref{fig::snr-10} and \ref{fig::snr20} plot the average PSNR over the CIFAR-10 test set as a function of the number of semantic pilots (from 1 to $10\,000$). 
In both cases, the aligners and DeepJSCCs are trained at the same SNR, i.e., $\snralign = \snrDeepJSCC$. In Figure \ref{fig::snr-10}, this common SNR is –10 dB, whereas in Figure \ref{fig::snr20} it is 20 dB. Each curve is averaged over five different random seeds, $\{42, 43, 44, 45, 46\}$, which induce distinct permutations of the training set. Across both SNR regimes, the convolutional equalizers (two-layer CNN and single-layer CNN) achieve high PSNR with only a few pilots—likely due to their architectural compatibility with the CNN-based DeepJSCC latent structure. In contrast, the linear equalizer requires substantially more pilots to reach similar fidelity, though it can outperform the CNN-based methods under heavy noise. Remarkably, the PFE—--despite being channel-agnostic—--attains around 30 dB PSNR with only $\sim\!10^3$ coefficients at high SNR, effectively compressing the signal without degrading quality.

Figures \ref{fig::psnr_snr_10} and \ref{fig::psnr_snr_20} plot the average PSNR at the receiver as a function of the inference SNR, which matches the $\snralign$. The two DeepJSCC systems were trained respectively at $\snrDeepJSCC = -10$ dB and $\snrDeepJSCC = 20$ dB. In each plot, the “no mismatch” curve represents the ideal case of a jointly trained encoder–decoder pair, while the remaining curves show performance with each semantic aligner compensating for both semantic and physical‐layer distortions.
\begin{figure*}[t]
    \centering
    \includegraphics[width=.84\textwidth, trim=0bp 0bp 0bp 0bp, clip]{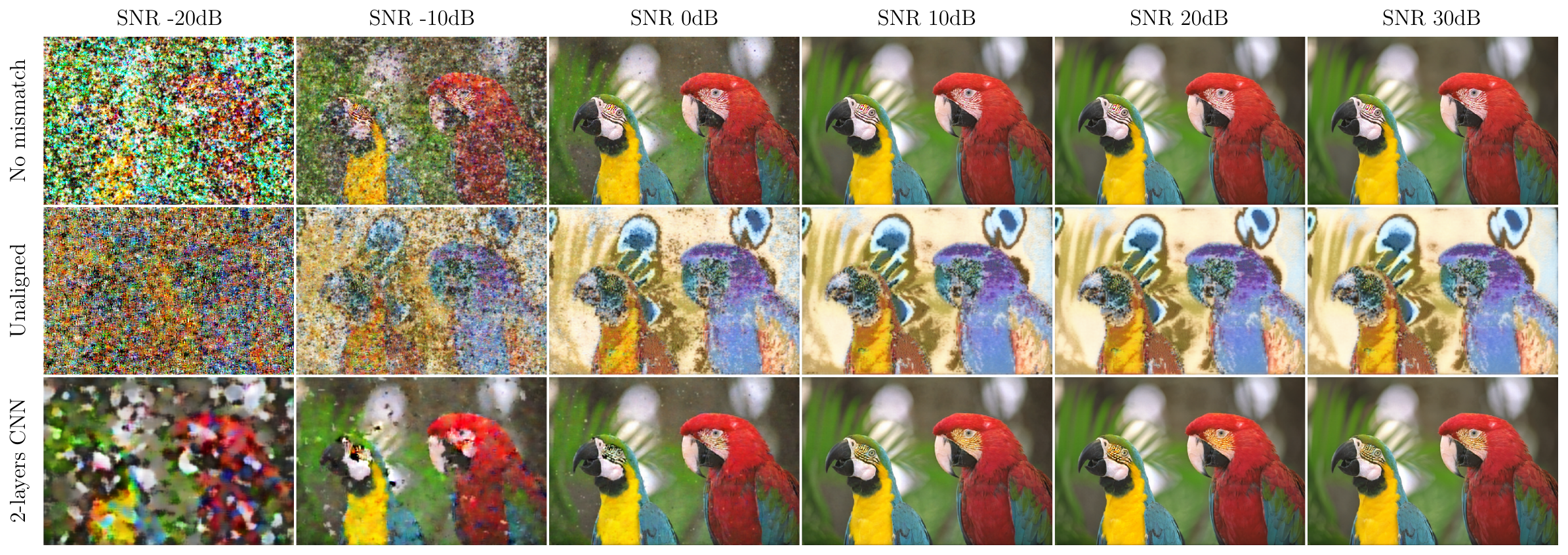}
    \caption{Reconstruction of an image from the Kodak Image Dataset under a fixed DeepJSCC trained solely under AWGN with SNR = 7 dB and a 2-layers CNN aligner trained on target SNRs \{-20, -10, 0, 10, 20, 30 \} dB over all the CIFAR-10 train set.}
    \label{fig::kodak_full}
\end{figure*}
When $\mathrm{SNR}_{\mathrm{Align}}$ is close to $\mathrm{SNR}_{\mathrm{DeepJSCC}}$, the no‐mismatch model achieves the highest PSNR, having been optimized around that operating point. However, as $\mathrm{SNR}_{\mathrm{Align}}$ deviates, the aligners progressively close the gap. In Fig. \ref{fig::psnr_snr_10}, for example, the two‐layer CNN aligner closely matches the no‐mismatch performance across a wide SNR range, demonstrating its ability to adapt to both semantic mismatches and varying channel conditions. Conversely, in Fig. \ref{fig::psnr_snr_20}, inference SNRs below 0 dB see all aligners--—except the channel‐agnostic PFE—--surpass the no‐mismatch baseline under heavy noise. 
In both Figures \ref{fig::psnr_snr_10} and \ref{fig::psnr_snr_20}, the introduction of fading (solid lines) leads to a systematic degradation in PSNR across all methods.

Finally, Fig. \ref{fig::kodak_full} illustrates the 2‐layers CNN aligner’s ability to handle full‐resolution images (768 × 512) without any cropping or resizing. Here, the DeepJSCC autoencoder was trained at 7 dB, and the CNN aligner was separately trained at various inference SNRs \{–20, –10, 0, 10, 20, 30\} dB using the entire CIFAR-10 training set. Because the CNN aligners are inherently resolution‐agnostic, they successfully align the unaltered Kodak image—--whose dimensions differ from the typical 96 × 96 input shape used in Fig. \ref{fig:visual_psnr}—--showing their flexibility to operate on arbitrarily sized inputs. These results confirm that semantic aligners represent a fundamental building block for heterogeneous DeepJSCC systems.

\section{Conclusions}
\label{sec:conclusions}

In this paper, we proposed a semantic equalization framework for DeepJSCC using three equalizer types: linear maps, lightweight neural networks, and zero-shot PFE. 
Our experiments on image reconstruction show that convolutional aligners rapidly achieve near-ideal quality with only a few semantic pilots, while the frame-based method offers robust, immediate deployment without retraining. The linear equalizer, in turn, proves most resilient under heavy noise but requires substantially more pilots to reach comparable fidelity. Moreover, linear, MLP, and PFE aligners are tied to a fixed input resolution and scale poorly with image size, whereas single- and two-layer CNN aligners are resolution-agnostic and maintain a constant parameter count. These insights clarify the trade-offs between robustness, data efficiency, computational complexity, and scalability, and provide practical guidance for integrating semantic equalization into DeepJSCC. Future work will extend these techniques to dynamic channel conditions, multi-carrier/antenna systems, and broader tasks such as speech, video, and text.

\begin{table}[t]
    \centering
    \caption{Studied aligners and their main characteristics.}
    \label{tab::aligners_characteristics}
    \renewcommand{\arraystretch}{1.2}
    \resizebox{\columnwidth}{!}{%
    \begin{tabular}{|lrcccc|}
        \hline
        \makecell{\textbf{Semantic}\\ \textbf{Aligners}} & 
        \textbf{\#Parameters} & 
        \makecell{\textbf{Receptive}\\ \textbf{Field}} & 
        \makecell{\textbf{Activation}\\ \textbf{Function}} & 
        \makecell{\textbf{Resolution}\\ \textbf{Agnostic}} &
        \makecell{\textbf{Bandwidth}\\ \textbf{Ratio}} \\
        \hline
        Linear       & 84,934,656  & All & ---    & No  & 1/6 \\
        MLP          & 169,887,745 & All & PReLU  & No  & 1/6 \\
        1-layer CNN  & 6,416       & 5   & ---    & Yes & 1/6 \\
        2-layer CNN  & 12,833      & 10  & PReLU  & Yes & 1/6 \\
        PFE          & ---         & All & ---    & No  & 2k/d \\
        \hline
    \end{tabular}
    }
\end{table}

\addtolength{\textheight}{-0.15in}
\balance
\bibliographystyle{IEEEtran}
\bibliography{refs}

\begin{thebibliography}{10}
\providecommand{\url}[1]{#1}
\csname url@samestyle\endcsname
\providecommand{\newblock}{\relax}
\providecommand{\bibinfo}[2]{#2}
\providecommand{\BIBentrySTDinterwordspacing}{\spaceskip=0pt\relax}
\providecommand{\BIBentryALTinterwordstretchfactor}{4}
\providecommand{\BIBentryALTinterwordspacing}{\spaceskip=\fontdimen2\font plus
\BIBentryALTinterwordstretchfactor\fontdimen3\font minus \fontdimen4\font\relax}
\providecommand{\BIBforeignlanguage}[2]{{%
\expandafter\ifx\csname l@#1\endcsname\relax
\typeout{** WARNING: IEEEtran.bst: No hyphenation pattern has been}%
\typeout{** loaded for the language `#1'. Using the pattern for}%
\typeout{** the default language instead.}%
\else
\language=\csname l@#1\endcsname
\fi
#2}}
\providecommand{\BIBdecl}{\relax}
\BIBdecl

\bibitem{shannon1998mathematical}
C.~E. Shannon and W.~Weaver, \emph{The mathematical theory of communication}.\hskip 1em plus 0.5em minus 0.4em\relax University of Illinois press, 1998.

\bibitem{gunduz2024joint}
D.~G{\"u}nd{\"u}z, M.~A. Wigger, T.-Y. Tung, P.~Zhang, and Y.~Xiao, ``Joint source--channel coding: Fundamentals and recent progress in practical designs,'' \emph{Proceedings of the IEEE}, 2024.

\bibitem{farsad2018text}
N.~Farsad, M.~Rao, and A.~Goldsmith, ``Deep learning for joint source-channel coding of text,'' in \emph{Proc. of IEEE ICASSP}, 2018, pp. 2326--2330.

\bibitem{bourtsoulatze2019deep}
E.~Bourtsoulatze, D.~B. Kurka, and D.~G{\"u}nd{\"u}z, ``Deep joint source-channel coding for wireless image transmission,'' \emph{IEEE Trans. on Cognitive Commun. and Networking}, vol.~5, no.~3, pp. 567--579, 2019.

\bibitem{erdemir2023generative}
E.~Erdemir, T.-Y. Tung, P.~L. Dragotti, and D.~G{\"u}nd{\"u}z, ``Generative joint source-channel coding for semantic image transmission,'' \emph{IEEE Journal on Selected Areas in Commun.}, vol.~41, no.~8, pp. 2645--2657, 2023.

\bibitem{bian2023space}
C.~Bian, Y.~Shao, H.~Wu, and D.~G{\"u}nd{\"u}z, ``Space-time design for deep joint source channel coding of images over mimo channels,'' in \emph{2023 IEEE 24th International Workshop on Signal Processing Advances in Wireless Communications (SPAWC)}.\hskip 1em plus 0.5em minus 0.4em\relax IEEE, 2023, pp. 616--620.

\bibitem{wu2024deep}
H.~Wu, Y.~Shao, C.~Bian, K.~Mikolajczyk, and D.~G{\"u}nd{\"u}z, ``Deep joint source-channel coding for adaptive image transmission over mimo channels,'' \emph{IEEE Transactions on Wireless Communications}, 2024.

\bibitem{letafati2024deep}
M.~Letafati, S.~A.~A. Kalkhoran, E.~Erdemir, B.~H. Khalaj, H.~Behroozi, and D.~G{\"u}nd{\"u}z, ``Deep joint source channel coding for privacy-aware end-to-end image transmission,'' \emph{arXiv preprint arXiv:2412.17110}, 2024.

\bibitem{kurka2020DJSCC}
D.~B. Kurka and D.~G{\"u}nd{\"u}z, ``Deepjscc-f: Deep joint source-channel coding of images with feedback,'' \emph{IEEE journal on selected areas in information theory}, vol.~1, no.~1, pp. 178--193, 2020.

\bibitem{xu2023deep}
J.~Xu, T.-Y. Tung, B.~Ai, W.~Chen, Y.~Sun, and D.~G{\"u}nd{\"u}z, ``Deep joint source-channel coding for semantic communications,'' \emph{IEEE communications Magazine}, vol.~61, no.~11, pp. 42--48, 2023.

\bibitem{strinati20216g}
E.~C. Strinati and S.~Barbarossa, ``{6G} networks: Beyond shannon towards semantic and goal-oriented communications,'' \emph{Computer Networks}, vol. 190, p. 107930, 2021.

\bibitem{strinati2024goal}
E.~C. Strinati, P.~Di~Lorenzo \emph{et~al.}, ``Goal-oriented and semantic communication in {6G} {AI}-native networks: The {6G-GOALS} approach,'' in \emph{Proc. of {EuCNC/6G} Summit)}, 2024, pp. 1--6.

\bibitem{luo2022semantic}
X.~Luo, H.-H. Chen, and Q.~Guo, ``Semantic communications: Overview, open issues, and future research directions,'' \emph{IEEE Wireless Communications}, vol.~29, no.~1, pp. 210--219, 2022.

\bibitem{getu2023semantic}
T.~M. Getu, G.~Kaddoum, and M.~Bennis, ``Semantic communication: A survey on research landscape, challenges, and future directions,'' \emph{Proceedings of the IEEE}, vol. 112, no.~11, pp. 1649--1685, 2024.

\bibitem{moschella2022relative}
L.~Moschella, V.~Maiorca, M.~Fumero, A.~Norelli, F.~Locatello, and E.~Rodol{\`a}, ``Relative representations enable zero-shot latent space communication,'' in \emph{Intern. Conf. on Learning Representations}, 2023.

\bibitem{fiorellino2024dynamic}
S.~Fiorellino, C.~Battiloro, E.~C. Strinati, and P.~Di~Lorenzo, ``Dynamic relative representations for goal-oriented semantic communications,'' in \emph{Proc. of EUSIPCO}, 2024, pp. 2107--2111.

\bibitem{maiorca2024latent}
V.~Maiorca, L.~Moschella, M.~Fumero, F.~Locatello, and E.~Rodol{\`a}, ``Latent space translation via inverse relative projection,'' \emph{arXiv preprint arXiv:2406.15057}, 2024.

\bibitem{fiorellino2025frame}
S.~Fiorellino, C.~Battiloro, E.~C. Strinati, and P.~Di~Lorenzo, ``Frame-based zero-shot semantic channel equalization for {AI}-native communications,'' \emph{arXiv preprint arXiv:2507.17835}, 2025.

\bibitem{huttebraucker2024relative}
T.~H{\"u}ttebr{\"a}ucker, S.~Fiorellino, M.~Sana, P.~Di~Lorenzo, and E.~C. Strinati, ``Relative representations of latent spaces enable efficient semantic channel equalization,'' in \emph{Proc. IEEE GLOBECOM}, 2024.

\bibitem{pandolfo2025mimo}
M.~E. Pandolfo, S.~Fiorellino, E.~C. Strinati, and P.~Di~Lorenzo, ``Latent space alignment for ai-native mimo semantic communications,'' in \emph{2025 International Joint Conference on Neural Networks}.\hskip 1em plus 0.5em minus 0.4em\relax IEEE, 2025.

\bibitem{huttebraucker2025ris}
T.~H{\"u}ttebr{\"a}ucker, M.~E. Pandolfo, S.~Fiorellino, E.~C. Strinati, and P.~Di~Lorenzo, ``Ris-aided latent space alignment for semantic channel equalization,'' \emph{arXiv preprint arXiv:2507.16450}, 2025.

\bibitem{krizhevsky2009learning}
A.~Krizhevsky, G.~Hinton \emph{et~al.}, ``Learning multiple layers of features from tiny images,'' 2009.

\bibitem{kokakimagedataset}
\BIBentryALTinterwordspacing
R.~Franzen, ``Kodak lossless true color image suite.'' [Online]. Available: \url{http://r0k.us/graphics/kodak/}
\BIBentrySTDinterwordspacing

\end{thebibliography}

\end{document}